\documentclass[a4,10pt,twocolumn]{article}%

\usepackage{graphicx}
\DeclareGraphicsExtensions{.pdf}
\usepackage{latexsym}
\usepackage{amssymb}
\usepackage{amsmath}
\usepackage{url}
\usepackage[paperwidth=210mm, paperheight=297mm, top=25mm, left=21mm, right=23mm, bottom=31mm]{geometry}


\newcommand{\C}{\mathbb{C}}
\newcommand{\R}{\mathbb{R}}
\newcommand{\be}{\begin{equation}}
\newcommand{\ee}{\end{equation}}

\newenvironment{inmargins}[1]{\begin{list}{}{
  \leftmargin=#1 \rightmargin=#1 \parsep=0pt
  \partopsep=0pt}\item[]}{\end{list}}

\title{Non-constant bounded holomorphic functions of hyperbolic numbers --
       Candidates for hyperbolic activation functions}

\begin{document}
\date{}

\twocolumn[
\vspace*{0pt}
\begin{center}
{\Large \bf Non-constant bounded holomorphic functions of hyperbolic numbers --
       Candidates for hyperbolic activation functions
\\[10pt]}
{\large * Eckhard Hitzer (University of Fukui)\\[18pt]}
\end{center}

\begin{inmargins}{15mm}
   {\bf Abstract-- } 
The Liouville theorem states that bounded holomorphic complex functions are necessarily constant. Holomorphic functions fulfill the socalled Cauchy-Riemann (CR) conditions. The CR conditions mean that a complex $z$-derivative is independent of the direction. Holomorphic functions are ideal for activation functions of complex neural networks, but the Liouville theorem makes them useless. Yet recently the use of hyperbolic numbers, lead to the construction of hyperbolic number neural networks. We will describe the Cauchy-Riemann conditions for hyperbolic numbers and show that there exists a new interesting type of bounded holomorphic functions of hyperbolic numbers, which are not constant. We give examples of such functions. They therefore substantially expand the available candidates for holomorphic activation functions for hyperbolic number neural networks. \\

\end{inmargins}
]

\section{Introduction}

For the sake of mathematical clarity, we first carefully review the notion of holomorphic functions in the two number systems of complex and hyperbolic numbers. 

The Liouville theorem states that bounded holomorphic complex functions 
$f: \C \rightarrow \C$ are necessarily constant \cite{KG:HolFunct}. Holomorphic functions are functions that fulfill the socalled Cauchy-Riemann (CR) conditions. The CR conditions mean that a complex $z$-derivative 
\be 
\frac{df(z)}{dz}, \,\,\, z=x+iy \in \C, \,\,\, x,y \in \R, \,\,\,ii=-1,
\ee 
is independent of the direction with respect to which the incremental ratio, that defines the derivative, is taken \cite{FC:MinkST}. Holomorphic functions would be ideal for activation functions of complex neural networks, but the Liouville theorem means that careful measures need to be taken in order to avoid poles (where the function becomes infinite). 

Yet recently the use of hyperbolic numbers 
\be 
  z=x+h\, y, \,\,\,h^2=1, \,\,\,
  x,y \in \R, \,\,\, h\notin \R. 
\ee 
lead to the construction of hyperbolic number neural networks. We will describe the generalized Cauchy-Riemann conditions for hyperbolic numbers and show that there exist bounded holomorphic functions of hyperbolic numbers, which are not constant. We give a new example of such a function. They are therefore excellent candidates for holomorphic activation functions for hyperbolic number neural networks \cite{BS:HypMP,NB:DecBHypN}. In \cite{NB:DecBHypN} it was shown, that hyperbolic number neural networks allow to control the angle of the decision boundaries (hyperplanes) of the real and the unipotent $h$-part of the output. But Buchholz argued in \cite{SB:thesis}, p. 114, that 
\begin{quote}
Contrary to the complex case, the hyperbolic logistic function is bounded. This
is due to the absence of singularities. Thus, in general terms, this seems to be a suitable activation function. Concretely, the following facts, however, might be of disadvantage. The real and imaginary part have different squashing values.
Both component functions do only significantly differ from zero around 
the lines\footnote{Note that we slightly correct the two formulas of Buchholz, because we think it necessary to delete $e_1$ in Buchholz' original $x = y e_1 (x > 0)$, etc. }
$x = y\,\, (x > 0)$ and $-x = y\,\, (x < 0)$.
\end{quote}

Complex numbers are isomorphic to the Clifford geometric algebra $Cl_{0,1}$ which is generated by a single vector $e_1$ of negative square $e_1=-1$, with algebraic basis $\{1, e_1\}$. The isomorphism $\C \cong Cl_{0,1}$ is realized by mapping $i \mapsto e_1$. 

Hyperbolic numbers are isomorphic to the Clifford geometric algebra $Cl_{1,0}$ which is generated by a single vector $e_1$ of positive square $e_1=+1$, with algebraic basis $\{1, e_1\}$. The isomorphism between hyperoblic numbers and $Cl_{1,0}$ is realized by mapping $h \mapsto e_1$.

\section{Complex variable functions}

We follow the treatment given in \cite{FC:MinkST}. 
We assume a complex function given by an absolute convergent power series. 
\be 
  w = f(z) = f(x+iy) = u(x,y) + i v(x,y),
\ee 
where $u,v: \R^2 \rightarrow \R$ are real functions of the real variables $x,y$. Since $u,v$ are obtained in an algebraic way from the complex number $z=x+iy$, they cannot be arbitrary functions but must satisfy certain conditions. There are several equivalent ways to obtain these conditions. Following Riemann, we state that a function $w = f(z) = u(x,y) + i v(x,y)$ is a function of the complex variable $z$ if its derivative is independent of the direction (in the complex plane) with respect to which the incremental ratio is taken. This requirement leads to two partial differential equations, named after Cauchy and Riemann (CR), which relate $u$ and $v$. 

One method for obtaining these equations is the following. We consider the expression $w = u(x,y) + i v(x,y)$ only as a function of $z$, but not of $\bar{z}$, i.e. the derivative with respect to $\bar{z}$ shall be zero. First we  perform the bijective substitution 
\be 
  x = \frac{1}{2}(z+\bar{z}), \qquad 
  y = -i\frac{1}{2}(z-\bar{z}),
  \label{eq:replacexy}
\ee 
based on $z=x+iy, \bar{z} = x-iy$. For computing the derivative $w_{,\bar{z}}=\frac{d w}{d\bar{z}}$ with the help of the chain rule we need the derivatives of $x$ and $y$ of \eqref{eq:replacexy}
\be 
  x_{,\bar{z}} = \frac{1}{2}, \qquad 
  y_{,\bar{z}} = \frac{1}{2}i.
\ee 
Using the chain rule we obtain
\begin{align} 
  w_{,\bar{z}} 
  &= u_{,x}x_{,\bar{z}} + u_{,y}y_{,\bar{z}}
    + i (v_{,x}x_{,\bar{z}} + v_{,y}y_{,\bar{z}})
  \nonumber \\ 
  &= \frac{1}{2}u_{,x} + \frac{1}{2}iu_{,y}
    + i (\frac{1}{2}v_{,x} + \frac{1}{2}iv_{,y})
  \nonumber \\ 
  &= \frac{1}{2}[u_{,x}-v_{,y} + i(v_{,x}+u_{,y} )]
  \stackrel{!}{=} 0. 
  \label{eq:rivanish}
\end{align} 
Requiring that both the real and the imaginary part of \eqref{eq:rivanish} vanish we obtain the \textit{Cauchy-Riemann conditions}
\be 
  u_{,x}=v_{,y}, \qquad u_{,y} = -v_{,x} .
  \label{eq:complexCR}
\ee 
Functions of a complex variable that fulfill the CR conditions are functions of $x$ and $y$, but they are only functions of $z$, not of $\bar{z}$. 

It follows from \eqref{eq:complexCR}, that both $u$ and $v$ fulfill the \textit{Laplace equation}
\be 
  u_{,xx} = v_{,yx} = v_{,xy} = -u_{,yy} 
  \,\,\Leftrightarrow \,\,
  u_{,xx}+u_{,yy}=0,
\ee 
and similarly 
\be 
  v_{,xx}+v_{,yy}=0.
\ee 
The Laplace equation is a simple example of an elliptic partial differential equation. The general theory of solutions to the Laplace equation is known as potential theory. The solutions of the Laplace equation are called \textit{harmonic} functions and are important in many fields of science, notably the fields of electromagnetism, astronomy, and fluid dynamics, because they can be used to accurately describe the behavior of electric, gravitational, and fluid potentials. In the study of heat conduction, the Laplace equation is the steady-state heat equation \cite{Wiki:LaplaceEqu}.

\textit{Liouville's theorem} \cite{KG:HolFunct} states, that any bounded holomorphic function $f:\C \rightarrow \C$, which fulfills the CR conditions is constant. Therefore for complex neural networks it is not very meaningful to use holomorphic functions as activation functions. 
If they are used, special measures need to be taken to avoid poles in the complex plane. 
Instead separate componentwise (split) real scalar functions for the real part $g_r: \R \rightarrow \R, u(x,y) \mapsto g_r(u(x,y))$, and for the imaginary part $g_i: \R \rightarrow \R, v(x,y) \mapsto g_i(v(x,y))$, are usually adopted. Therefore a standard split activation function in the complex domain is given by 
\be 
  g(u(x,y)+iv(x,y))
  = g_r(u(x,y)) + i g_i(v(x,y)) .
\ee

\section{Hyperbolic numbers}

Hyperbolic numbers are also known as \textit{split-complex} numbers. They form a two-dimensional commutative algebra. 
The canonical hyperbolic system of numbers is defined \cite{FC:MinkST} by 
\be 
  z=x+h\, y, \,\,\,h^2=1, \,\,\,
  x,y \in \R, \,\,\, h\notin \R. 
\ee 

The \textit{hyperbolic conjugate} is defined as
\be 
  \bar{z}=x-h\, y .
\ee 
Taking the hyperbolic conjugate corresponds in the isomorphic algebra $Cl_{1,0}$ to taking the main involution (grade involution), which maps $1\mapsto 1, e_1 \mapsto -e_1$. 

The \textit{hyperbolic invariant} (corresponding to the Lorentz invariant in physics for $y = ct$), or \textit{modulus}, is defined as
\be 
  z\bar{z} = (x+h\, y)(x-h\, y) = x^2 - y^2,
\ee 
which is not positive definite. 

Hyperbolic numbers are fundamentally different from complex numbers. Complex numbers and quaternions are division algebras, every non-zero element has a unique inverse. Hyperbolic numbers do not always have an inverse, but instead there are idempotents and divisors of zero. 

We can define the following \textit{idempotent basis}
\be 
  n_1 = \frac{1}{2}(1+h), \qquad
  n_2 = \frac{1}{2}(1-h), 
\ee 
which fulfills 
\begin{align} 
  n_1^2 &= \frac{1}{4}(1+h)(1+h)
        = \frac{1}{4}(2+2h)
        = n_1,  
  \nonumber \\
  n_2^2 & = n_2, \qquad  n_1+n_2 = 1,
  \nonumber \\
  n_1n_2 &= \frac{1}{4}(1+h)(1-h)
  = \frac{1}{4}(1-1)=0,
  \nonumber \\
  \bar{n}_1 &= n_2, \quad \bar{n}_2 = n_1.
  \label{eq:idbasis}
\end{align}
The inverse basis transformation is simply
\be 
  1 = n_1+n_2, \qquad 
  h = n_1-n_2 . 
\ee 
Setting
\be 
  z = x+ h y = \xi n_1 + \eta n_2,
\ee 
we get the corresponding coordinate transformation
\be 
  x = \frac{1}{2}(\xi+ \eta), \qquad 
  y = \frac{1}{2}(\xi- \eta),
\ee 
as well as the inverse coordinate transformation
\be 
  \xi  = x+y \in \R, \qquad 
  \eta = x-y \in \R. 
\ee 
The hyperbolic conjugate becomes, due to \eqref{eq:idbasis}, in the idempotent basis
\be 
  \bar{z} =  \xi\bar{n}_1 + \eta \bar{n}_2
  = \eta n_1 +  \xi n_2 .
  \label{eq:idconj}
\ee 
In the idempotent basis, using \eqref{eq:idconj} and \eqref{eq:idbasis}, the hyperbolic invariant becomes multiplicative
\begin{align}
  z\bar{z} &=  (\xi n_1 + \eta n_2)(\eta n_1 +  \xi n_2)
  \nonumber \\ 
  &= \xi\eta (n_1+n_2) = \xi\eta = x^2-y^2. 
\end{align}
In the following we consider the product and quotient of two hyperbolic numbers $z,z'$ both expressed in the idempotent basis $\{n_1, n_2 \}$
\be 
  zz' = (\xi n_1 + \eta n_2)(\xi{}' n_1 + \eta{}' n_2) 
  = \xi\xi{}' n_1 + \eta\eta{}' n_2,
\ee 
and 
\begin{align}
  \frac{z}{z'} 
  &= \frac{\xi n_1 + \eta n_2}{\xi{}' n_1 + \eta{}' n_2}
  = \frac{z\bar{z}'}{z'\bar{z}'}
 \nonumber \\
  &= \frac{(\xi n_1 + \eta n_2)(\eta{}' n_1 +  \xi{}' n_2)}%
    {(\xi{}' n_1 + \eta{}' n_2)(\eta{}' n_1 +  \xi{}' n_2)}
 \nonumber \\
  &=  \frac{(\xi\eta{}' n_1 + \eta\xi{}' n_2)(\eta{}' n_1 +  \xi{}' n_2)}%
    {\xi{}'\eta{}'}
 \nonumber \\
  &= \frac{\xi}{\xi{}'} n_1 +\frac{\eta}{\eta{}'} n_2 . 
  \label{eq:divzero}
\end{align}
Because of \eqref{eq:divzero} it is not possible to divide by $z'$ if $\xi{}'=0$, or if $\eta{}'=0$. Moreover, the product of a hyperbolic number with $\xi{}=0$ (on the $n_2$ axis) times a hyperbolic number with $\eta=0$ (on the $n_1$ axis) is 
\be 
  (\xi n_1 + 0 n_2)(0n_1 + \eta n_2) 
  = \xi\eta n_1n_2 
  = 0,
  \label{eq:prdivzero}
\ee 
due to \eqref{eq:idbasis}. We repeat that in \eqref{eq:prdivzero} the product is zero, even though the factors are non-zero. The numbers $\xi n_1, \eta n_2$ along the $n_1, n_2$ axis are therefore called \textit{divisors of zero}. The divisors of zero have no inverse. 

\begin{figure}
\begin{center}
\includegraphics[width=0.45\textwidth]{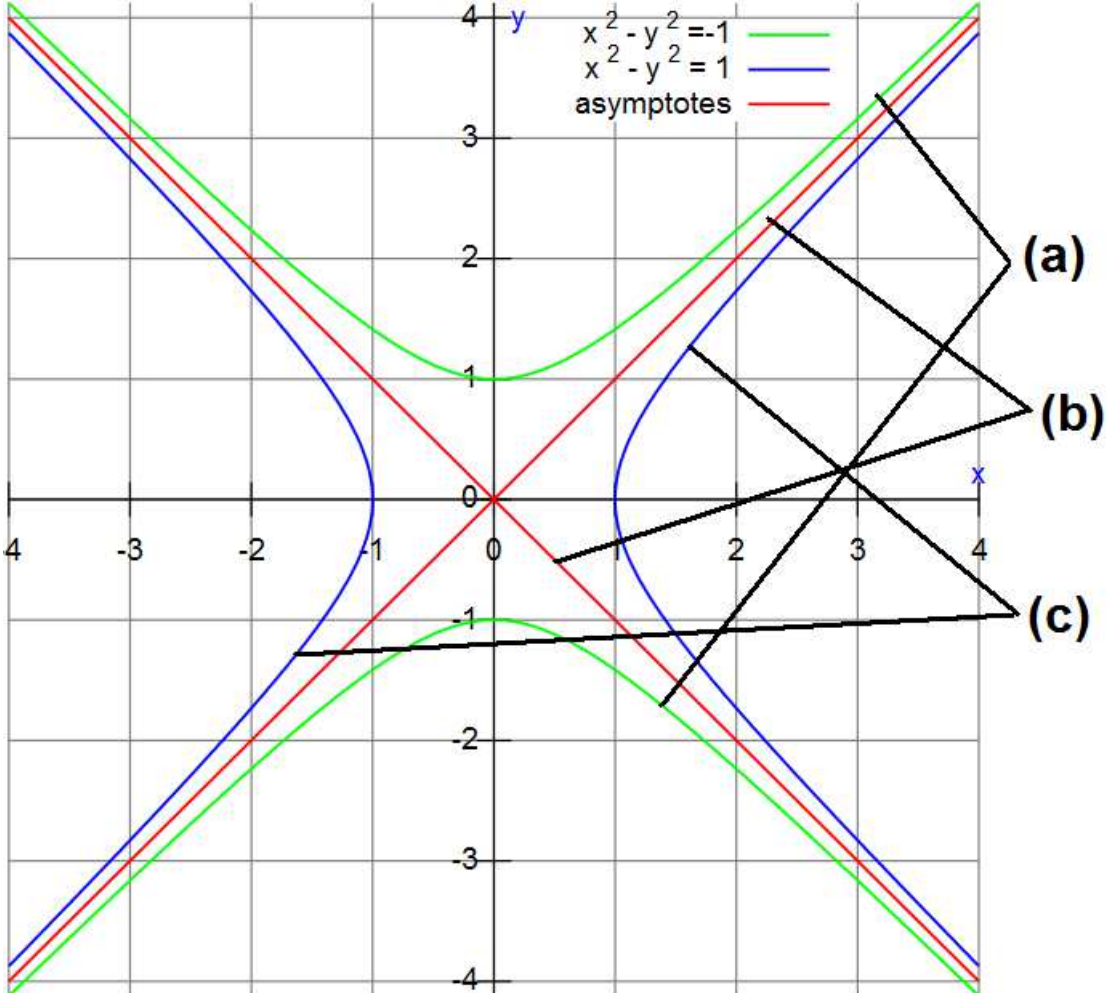}
\end{center}
\caption{The hyperbolic number plane \cite{Wiki:HypPlane} with horizontal $x$-axis and vertical $yh$-axis, showing: 
(a) Hyperbolas with modulus $z\bar{z}=-1$ (green).
(b) Straight lines with modulus $z\bar{z}=0 \Leftrightarrow x^2=y^2$ (red), i.e. divisors of zero. 
(c) Hyperbolas with modulus $z\bar{z}=1$ (blue).
\label{fg:HypPlane}}
\end{figure}

The hyperbolic plane with the diagonal lines of 
divisors of zero (b), and the pairs of hyperbolas with constant modulus 
$z\bar{z}=1$ (c), 
and $z\bar{z}=-1$ (a) is shown in Fig. \ref{fg:HypPlane}.

\section{Hyperbolic number functions}

We assume a hyperbolic number function given by an absolute convergent power series 
\begin{align}
  w &= f(z) = f(x+hy) = u(x,y) + h v(x,y), 
  \nonumber \\
  h^2 &= 1, \quad h\notin \R. 
\end{align}
where $u,v: \R^2 \rightarrow \R$ are real functions of the real variables $x,y$.
An example of a hyperbolic number function is the exponential function
\begin{align}
  e^{z} &= e^{x+hy}= e^x e^{hy} = e^x (\cosh y + h \sinh y) 
  \nonumber \\
  &= u(x,y) + h v(x,y), 
  \label{eq:hypexp}
\end{align} 
with 
\be 
  u(x,y) = e^x \cosh y , \qquad 
  v(x,y) = e^x \sinh y .
\ee 

Since $u,v$ are obtained in an algebraic way from the hyperbolic number $z=x+hy$, they cannot be arbitrary functions but must satisfy certain conditions. There are several equivalent ways to obtain these conditions. A function $w = f(z) = u(x,y) + h v(x,y)$ is a function of the hyperbolic variable $z$, if its derivative is independent of the direction (in the hyperbolic plane) with respect to which the incremental ratio is taken. This requirement leads to two partial differential equations, so called \textit{generalized Cauchy-Riemann} (GCR) conditions, which relate $u$ and $v$. 

To obtain the GCR conditions we consider the expression $w = u(x,y) + h v(x,y)$ only as a function of $z$, but not of $\bar{z}=x-hy$, i.e. the derivative with respect to $\bar{z}$ shall be zero. First we  perform the bijective substitution 
\be 
  x = \frac{1}{2}(z+\bar{z}), \qquad 
  y = h\frac{1}{2}(z-\bar{z}),
  \label{eq:hreplacexy}
\ee 
based on $z=x+hy, \bar{z} = x-hy$. For computing the derivative $w_{,\bar{z}}=\frac{d w}{d\bar{z}}$ with the help of the chain rule  we need the derivatives of $x$ and $y$ of \eqref{eq:hreplacexy}
\be 
  x_{,\bar{z}} = \frac{1}{2}, \qquad 
  y_{,\bar{z}} = -\frac{1}{2}h.
\ee 
Using the chain rule we obtain
\begin{align} 
  w_{,\bar{z}} 
  &= u_{,x}x_{,\bar{z}} + u_{,y}y_{,\bar{z}}
    + h (v_{,x}x_{,\bar{z}} + v_{,y}y_{,\bar{z}})
  \nonumber \\ 
  &= \frac{1}{2}u_{,x} - \frac{1}{2}hu_{,y}
    + h (\frac{1}{2}v_{,x} - \frac{1}{2}hv_{,y})
  \nonumber \\ 
  &= \frac{1}{2}[u_{,x}-v_{,y} + h ( v_{,x} - u_{,y} )]
  \stackrel{!}{=} 0. 
  \label{eq:hrivanish}
\end{align} 
Requiring that both the real and the $h$-part of \eqref{eq:hrivanish} vanish we obtain the GCR conditions
\be 
  u_{,x} = v_{,y}, \qquad u_{,y} = v_{,x} .
  \label{eq:GCR}
\ee 
Functions of a hyperbolic variable that fulfill the GCR conditions are functions of $x$ and $y$, but they are only functions of $z$, not of $\bar{z}$. 
Such functions are called (hyperbolic) holomorphic functions.

It follows from \eqref{eq:GCR}, that $u$ and $v$ fulfill the wave equation
\be 
  u_{,xx} = v_{,yx} = v_{,xy} = u_{,yy} \Leftrightarrow 
  u_{,xx}-u_{,yy}=0,
\ee 
and similarly 
\be 
  v_{,xx}-v_{,yy}=0.
\ee 

The wave equation is an important second-order linear partial differential equation for the description of waves -- as they occur in physics -- such as sound waves, light waves and water waves. It arises in fields like acoustics, electromagnetics, and fluid dynamics. The wave equation is the prototype of a hyperbolic partial differential equation \cite{Wiki:WaveEqu}.

Let us compute the partial derivatives $u_{,x}, u_{,y}$, $v_{,x}, v_{,y}$ for the exponential function $e^z$ of \eqref{eq:hypexp}: 
\begin{align}
  u_{,x} &= e^x \cosh y, \quad 
  u_{,y} = e^x \sinh y,
  \nonumber \\
  v_{,x} &= e^x \sinh y = u_{,y}, \quad
  v_{,y} = e^x \cosh y = u_{,x}.
  \label{eq:hypexpder}
\end{align}
We clearly see that the partial derivatives \eqref{eq:hypexpder} fulfill the GCR conditions \eqref{eq:GCR} for the exponential function $e^z$, as expected by its definition \eqref{eq:hypexp}. The exponential function $e^z$ is therefore a manifestly holomorphic hyperpolic function, but it is not bounded.

\begin{figure}
\begin{center}
\includegraphics[width=0.4\textwidth]{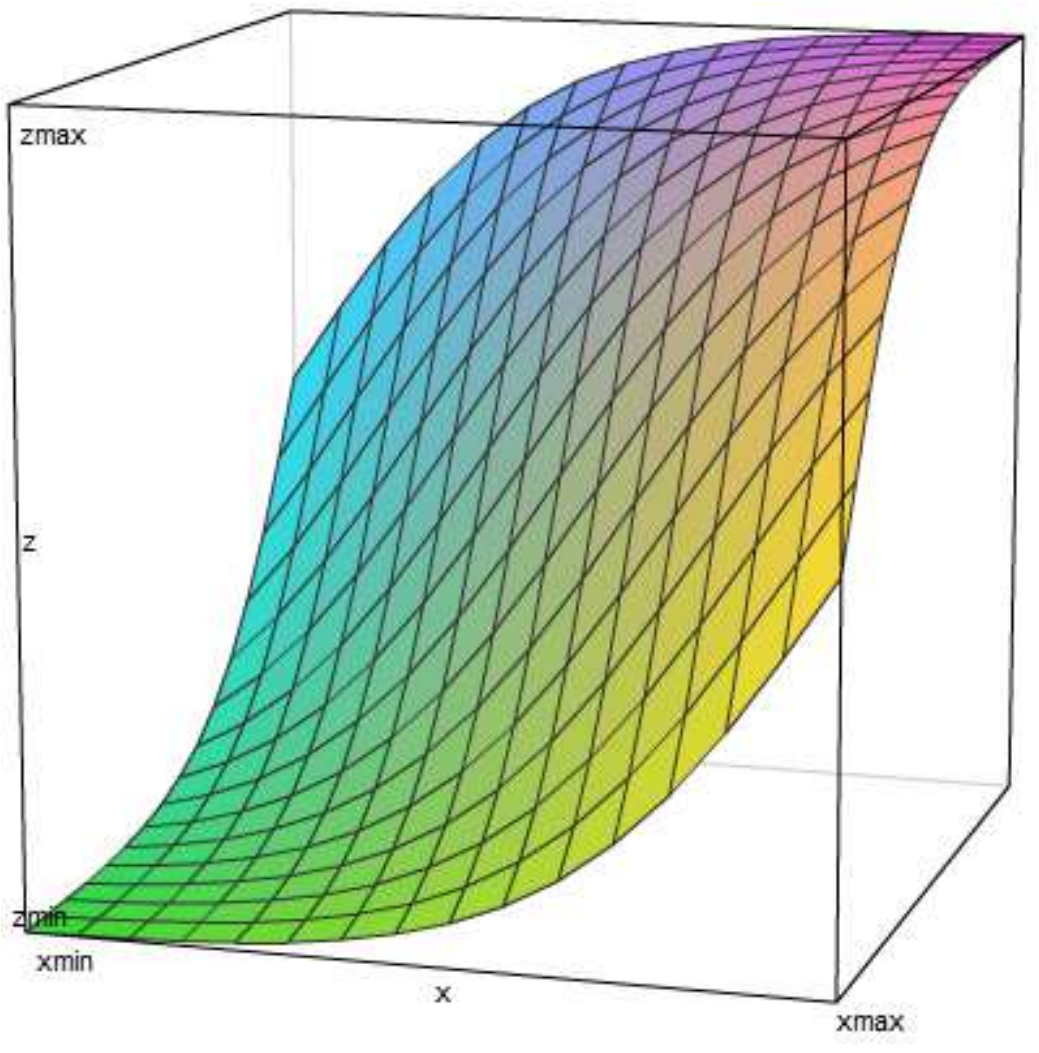}
\end{center}
\caption{Function $u(x,y) = {1}/({1+ e^{-x}e^{-y}})$. 
Horizontal axis $-3\leq x \leq 3$, 
from left corner into paper plane  $-3\leq y \leq 3$. 
Vertical axis $0\leq u \leq 1$. (Figure produced with \cite{3DGrapher}.)
\label{fg:fc1}}
\end{figure}

In the case of holomorphic hyperbolic functions the GCR conditions do not imply a Liouville type theorem like for holomorphic complex functions. This can most easily be demonstrated with a counter example
\begin{align}
  f(z) &= u(x,y) + h\, v(x,y), 
  \nonumber \\
  u(x,y) &= v(x,y) = \frac{1}{1+ e^{-x}e^{-y}} .
  \label{eq:ex1}
\end{align}
The function $u(x,y)$ is pictured in Fig. \ref{fg:fc1}.

Let us verify that the function $f$ of \eqref{eq:ex1} fulfills the GCR conditions
\begin{align}
  u_{,x} &=  \frac{-1}{(1+ e^{-x}e^{-y})^2} (-e^{-x}e^{-y})
  \nonumber \\
         &=  \frac{e^{-x}e^{-y}}{(1+ e^{-x}e^{-y})^2},
\end{align}
where we repeatedly applied the chain rule for differentiation. Similarly we obtain 
\be 
  u_{,y} = v_{,x} = v_{,y} = \frac{e^{-x}e^{-y}}{(1+ e^{-x}e^{-y})^2}.
\ee 
The GCR conditions \eqref{eq:GCR} are therefore clearly fulfilled, which means that the hyperbolic function $f(z)$ of \eqref{eq:ex1} is holomorphic. Since the exponential function $e^{-x}$ has a range of $(0,\infty)$, the product 
 $e^{-x} e^{-y}$ also has values in the range of $(0,\infty)$. Therefore the function  $1+e^{-x} e^{-y}$ has values in $(1,\infty)$, and the components of the function $f(z)$ of \eqref{eq:ex1} have values
\be 
  0 < \frac{1}{1+ e^{-x}e^{-y}} < 1 .
\ee 
We especially have 
\be 
  \lim_{x,y\rightarrow -\infty} \,\,\frac{1}{1+ e^{-x}e^{-y}}
  = 0,
\ee 
and 
\be 
  \lim_{x,y\rightarrow \infty} \,\,\frac{1}{1+ e^{-x}e^{-y}}
  = 1.
\ee

The function \eqref{eq:ex1} is \textit{representative} for how to turn any real neural node activation function $r(x)$ into holomorphic hyperbolic activation function via
\be 
  f(x) = r(x+y)\,(1+h).
\ee 

We note that in \cite{NB:DecBHypN,SB:thesis} another holomorphic hyperbolic activation function was studied, namely
\be 
  f'(z) = \frac{1}{1+e^{-z}} , 
\ee 
but compare the quote from \cite{SB:thesis}, p. 114, given in the introduction. The split activation function used in \cite{BS:HypMP}
\be 
  f''(x,y) = \frac{1}{1+e^{-x}} + h \frac{1}{1+e^{-y}},
\ee 
is clearly not holomorphic, because the real part $u= {1}/({1+e^{-x}})$ depends only on $x$ and not on $y$, and the $h$-part $v={1}/({1+e^{-y}})$ depends only on $y$ and not on $x$, thus the GCR conditions \eqref{eq:GCR} can not be fulfilled.

\section{Geometric interpretation of multiplication of hyperbolic numbers}

In order to geometrically interpret the product of two complex numbers, it proves useful to introduce polar coordinates in the complex plane. Similarly, for the geometric interpretation of the product of two hyperbolic numbers, we first introduce \textit{hyperbolic polar coordinates} for 
$z = x+hy $ with radial coordinate 
\be 
  \rho = \sqrt{|z\bar{z}|} = \sqrt{|x^2-y^2|}\, .
\ee
The hyperbolic polar coordinate transformation \cite{FC:MinkST} is then given as 
\begin{enumerate}
\item $x^2 > y^2$, $x > 0$:
$$
  \theta = \mathrm{artanh} \,(y/x), \qquad 
  z = \rho e^{h\theta},  
$$
i.e. the quadrant in the hyperbolic plane of Fig. \ref{fg:HypPlane} limitted by the diagonal idempotent lines, and including the positive $x$-axis (to the right).
\item $x^2 > y^2$, $x < 0$:
$$
  \theta = \mathrm{artanh} \,(y/x), \qquad
  z = -\rho e^{h\theta},  
$$
i.e. the quadrant in Fig. \ref{fg:HypPlane} including the negative $x$-axis (to the left).
\item $x^2 < y^2$, $y > 0$:
$$
  \theta = \mathrm{artanh} \,(x/y), \qquad
  z = h\rho e^{h\theta},
$$
i.e. the quadrant in Fig. \ref{fg:HypPlane} including the positive $y$-axis (top).
\item $x^2 < y^2$, $y < 0$:
$$
  \theta = \mathrm{artanh} \,(x/y), \qquad
  z = -h\rho e^{h\theta},
$$
i.e. the quadrant in Fig. \ref{fg:HypPlane} including the negative $y$-axis (bottom).
\end{enumerate}

The product of a constant hyperbolic number (assuming $a_x^2>a_y^2, a_x > 0$)
\begin{align}
  a &= a_x + h a_y = \rho_a e^{h\theta_a},
  \nonumber \\
  \rho_a &= \sqrt{a_x^2-a_y^2}, \qquad
  \theta_a = \mathrm{artanh} \,(a_y/a_x),
\end{align}
with a hyperbolic number $z$ (assuming $x^2 > y^2$, $x > 0$) in hyperbolic polar coordinates is
\be 
  az = \rho_a \,e^{h\theta_a}\,\rho\, e^{h\theta}
  = \rho_a\rho \,e^{h(\theta+\theta_a)} .
\ee 
The geometric interpretation is a scaling of the modulus $\rho \rightarrow \rho_a\rho$ and a hyperbolic rotation (movement along a hyperbola) 
$\theta \rightarrow \theta+\theta_a$. 

In the physics of Einstein's special relativistic space-time \cite{DL:GAfPh,EH:RPAppGA}, the hyperbolic rotation $\theta \rightarrow \theta+\theta_a$ corresponds to a Lorentz transformation from one inertial frame with constant velocity $\tanh \theta$ to another inertial frame with constant velocity $\tanh (\theta+\theta_a)$. 
Neural networks based on hyperbolic numbers (dimensionally extended to four-dimensional space-time) should therefore be ideal to compute with electromagnetic signals, including satellite transmission.

\section{Conclusion}

We have compared complex numbers and hyperbolic numbers, as well as complex functions and hyperbolic functions. We saw that according to Liouville's theorem bounded complex holomorphic functions are necessarily constant, \textit{but} non-constant bounded hyperbolic holomorphic functions exist. 
One such function has already beeng studied in \cite{NB:DecBHypN,SB:thesis}. We have studied a promising
example of a hyperbolic holomorphic function 
\be 
  f(z) = \frac{1+h}{1+e^{-x-y}}\,,
\ee 
in some detail. 
The distinct notions of idempotents and divisors of zero, special to hyperbolic numbers, were introduced. 
After further introducing hyperbolic polar coordinates, a geometric interpretation of the hyperbolic number multiplication was given.  

Hyerbolic neural networks offer, compared to complex neural networks, therefore the advantage of \textit{suitable bounded non-constant hyperbolic holomorphic activation functions}. It would certainly be of interest to study convergence, accuracy and decision boundaries of hyperbolic neural networks with the activation function \eqref{eq:ex1}, similar to \cite{NB:DecBHypN,SB:thesis}. 

\section*{Acknowledgment}

I want to acknowledge God \cite{NIV:John1}:
\begin{quote}
In the beginning was the Word\footnote{Greek term: \textit{logos}. Note: A Greek philosopher named Heraclitus first used the term Logos around 600 B.C. to designate the divine reason or plan which coordinates a changing universe.~\cite{BLB:Strong}}, and the Word was with God, and the Word was God. He was with God in the beginning. Through him all things were made; without him nothing was made that has been made. In him was life, and that life was the light of all mankind. 
\end{quote}
I want to thank my dear family, as well as T. Nitta and Y. Kuroe.


\begin{thebibliography}{99}

\bibitem{KG:HolFunct}
K. Guerlebeck et al, 
Holomorphic Functions in the Plane and n-dimensional Space, 
Birkhauser, 2008, chp. 7.3.3. 

\bibitem{BS:HypMP}
S. Buchholz, G. Sommer,
\textit{A hyperbolic multilayer perceptron},
Proceedings of the International Joint Conference on Neural Networks, Como,
Italy, vol. 2, 129/133 (2000).

\bibitem{NB:DecBHypN}
T. Nitta, S. Buchholz,
\textit{On the Decision Boundaries of Hyperbolic Neurons}, 
Proceedings of the International Joint Conference on Neural Networks, IJCNN'08-HongKong, June 1-6, 
2973/2979(2008). 

\bibitem{SB:thesis}
S. Buchholz, PhD Thesis, 
\textit{A Theory of Neural Computation with Clifford Algebras},
University of Kiel, 2005.

\bibitem{FC:MinkST}
F. Catoni et al, 
The Mathematics of Minkowski Space-Time, 
Birkhauser, 2008. 

\bibitem{Wiki:LaplaceEqu}
Laplace's equation, Wikipedia, accessed 24 August 2011, 
\url{http://en.wikipedia.org/wiki/Laplace's_equation}

\bibitem{Wiki:WaveEqu}
Wave equation, Wikipedia, accessed 24 August 2011, 
\url{http://en.wikipedia.org/wiki/Wave_equation}

\bibitem{3DGrapher}
Online 3D function grapher,
\url{http://www.livephysics.com/ptools/online-3d-function-grapher.php?}

\bibitem{Wiki:HypPlane}
Split-complex number, Wikipedia, accessed 29 August 2011, 
\url{http://en.wikipedia.org/wiki/Split-complex_number}

\bibitem{HI:Notes}
Notes of collaboration with H. Ishi, 
Feb. 2011, p. 15. 

\bibitem{DL:GAfPh}
C. Doran and A. Lasenby, 
\textit{Geometric Algebra for Physicists}, 
Cambridge University Press, Cambridge (UK), 2003.

\bibitem{EH:RPAppGA}
E. Hitzer, 
\textit{Relativistic Physics as Application of Geometric Algebra}, 
in K. Adhav (ed.), 
Proceedings of the International Conference on Relativity 2005 (ICR2005), University of Amravati, India, January 2005, 71/90(2005).

\bibitem{NIV:John1}
The Bible, New International Version (NIV), The Gospel according to John, chapter 1, verses 1-4, 
\url{http://www.biblegateway.com/}

\bibitem{BLB:Strong}
Strong's Bible lexicon entry G3056 for \textit{logos}, available online at Blue Letter Bible. 
\url{http://www.blueletterbible.org/lang/lexicon/lexicon.cfm?Strongs=G3056&t=KJV}


\end{thebibliography}
\end{document}